\documentclass[conference]{IEEEtran}
\IEEEoverridecommandlockouts
\usepackage{cite}
\usepackage{amsmath,amssymb,amsfonts}
\usepackage{algorithmic}
\usepackage{graphicx}
\usepackage{textcomp}
\usepackage{xcolor}
\def\BibTeX{{\rm B\kern-.05em{\sc i\kern-.025em b}\kern-.08em
  T\kern-.1667em\lower.7ex\hbox{E}\kern-.125emX}}
\usepackage[colorlinks,linkcolor=black,citecolor=black]{hyperref}
 
\begin{document}
\title{Arrhythmia Classifier using Binarized Convolutional Neural Network for Resource-Constrained Devices}

\author{\IEEEauthorblockN{1\textsuperscript{st} Ao Wang}
\IEEEauthorblockA{\textit{School of Electronic Science and}\\ \textit{Engineering} \\
\textit{Southeast University}\\
Nanjing, Jiangsu, China\\
Ao0323@outlook.com}
\and
\IEEEauthorblockN{2\textsuperscript{nd} Wenxing Xu}
\IEEEauthorblockA{\textit{School of Computer Science and}\\ \textit{Information Engineering} \\
\textit{Hefei University of Technology}\\
Hefei, Anhui, China \\
meaple.sfky@icloud.com}
\and
\IEEEauthorblockN{3\textsuperscript{rd} Hanshi Sun}
\IEEEauthorblockA{\textit{School of Electronic Science and}\\ \textit{Engineering} \\
\textit{Southeast University}\\
Nanjing, Jiangsu, China\\
preminstrel@gmail.com}
\and
\IEEEauthorblockN{4\textsuperscript{th} Ninghao Pu}
\IEEEauthorblockA{\textit{School of Electronic Science and}\\ \textit{Engineering} \\
\textit{Southeast University}\\
Nanjing, Jiangsu, China\\
1742237531@qq.com}
\and
\IEEEauthorblockN{5\textsuperscript{th} Zijin Liu}
\IEEEauthorblockA{\textit{School of Electronic Science and}\\ \textit{Engineering} \\
\textit{Southeast University}\\
Nanjing, Jiangsu, China\\
yzliuzijin@163.com}
\and
\IEEEauthorblockN{6\textsuperscript{th} Hao Liu\IEEEauthorrefmark{2}}
\IEEEauthorblockA{\textit{School of Electronic Science and}\\ \textit{Engineering} \\
\textit{Southeast University}\\
Nanjing, Jiangsu, China\\
nicky\_lh@seu.edu.cn}
}

\maketitle

\begin{abstract}
Monitoring electrocardiogram signals is of great significance for the diagnosis of arrhythmias. In recent years, deep learning and convolutional neural networks have been widely used in the classification of cardiac arrhythmias. However, the existing neural network applied to ECG signal detection usually requires a lot of computing resources, which is not friendlyF to resource-constrained equipment, and it is difficult to realize real-time monitoring. In this paper, a binarized convolutional neural network suitable for ECG monitoring is proposed, which is hardware-friendly and more suitable for use in resource-constrained wearable devices. Targeting the MIT-BIH arrhythmia database, the classifier based on this network reached an accuracy of 95.67\% in the five-class test. Compared with the proposed baseline full-precision network with an accuracy of 96.45\%, it is only 0.78\% lower. Importantly, it achieves 12.65 times the computing speedup, 24.8 times the storage compression ratio, and only requires a quarter of the memory overhead.
\end{abstract}


\section{Introduction}
Arrhythmia is a common disorder that has a considerable impact on human body health, causing morbidity alone or accompanied by other cardiovascular diseases\cite{b1}. A large number of people die of sudden death due to arrhythmia every year, so it is of great significance to identify arrhythmia as soon as possible. The recognition of electrocardiogram (ECG) is the most basic and simple method to diagnose arrhythmias for ECG contains the basic information about the state of the heart. However,  it will be a lot of wastage of medical resources if ECG signal recognition only relies on related experts or doctors to achieve. And the development of modern computers and information technology makes the identification of ECG signals more convenient. Due to the abundant information contained in ECG signals, accurate analysis of ECG signals is an undoubtedly complex task, which affects the diagnostic results.

Recent years have witnessed the unprecedented success that deep neural network(DNN) has achieved in the fields of speech recognition, image recognition and medical applications, and rhythm recognition using convolution neural network(CNN) has also attracted widespread attention\cite{b2, b3}. Though it’s of advantages for CNN to do rhythm recognition tasks, the deployment of CNN models to wearable devices with low computing resources and small memory space is still limited for the derivation of existing CNN model requires a lot of energy. As a computation-intensive and memory-intensive model, CNN's inference process contains many computations and memory-access operations, which imposes a huge power burden on wearable devices with limited hardware resources. Binarized Neural Networks(BNN) have attracted much attention cause of their lower memory consumption and higher computing speed, but it tends to bring a serious drop in terms of accuracy.

To balance the accuracy of ECG recognition and the overhead of hardware resources, we considered both effective network architecture and means of binarizing the network to reduce the memory overhead while maintaining considerable accuracy. In this study, we explored methods suitable for binarizing 1-D convolutional neural networks and adopted them to the proposed baseline DNN classifier, implementing a five-category binarized classifier for ECG signals. The contribution of this paper has the following three aspects:

\begin{itemize}
	\item A binarization scheme of CNN models suitable for 1-D ECG signal is proposed, which requires only a small memory overhead and has a high computational speed.
	\item It introduces a topographic-based strategy for adjusting the learning rate of the binarized network, which has a great impact on the accuracy of the binarized model.
	\item We further implement the binarization method and achieve an accuracy of 95.67\% in the five-class test, which is only 0.78\% lower than that of the baseline full-precision network\footnote{Code is available on: https://github.com/Christyao-ada/ECG-Bianrized}.
\end{itemize}

Experiments show that the proposed binarized model has considerable recognition performance, yet it achieves a $3.78\times$ reduction in runtime memory overhead, a $24.8\times$ reduction in storage footprint for weight parameters, and a $12.65\times$ runtime speedup. Compared with other state-of-the-arts, the proposed model is more suitable for deployment on resource-constrained wearable devices, making real-time ECG monitoring possible.

The remainder of this article is organized as follows. The background and some related works ever been done will be presented in \ref{related}. \ref{method} introduces the methods we use in the experiment. \ref{exp} shows the results of experiment and comparisons with other models. Finally, we conclude the work of this paper and look forward to future works in \ref{conclu}.

\section{Related Work}\label{related}
\subsection{Traditional pattern recognition methods applied to ECG}
The automatic diagnosis of arrhythmia can be realized by pattern recognition\cite{b3,b4}, and the support vector machine(SVM) system can also be used for heartbeat recognition and classification\cite{b5, b6, b7}. These methods are based on manually transforming the input into identifiable features, and the extraction of features is very dependent on labor resources. 

\subsection{Deep Learning methods applied to ECG recognition}
Recently, there has been an increasing amount of literature on ECG recognition applying deep learning methods. Feature extraction is learned autonomously from the model in the Artificial Neural Network(ANN), and deep learning can directly infer the type of arrhythmia from the original ECG signal. For instance, the PhysioNet Challenge proposed by G. D. Clifford et al. has successfully compared the four classifications of short single-lead ECG signals\cite{b8}. X. Fan et al. used MS-CNN to effectively extract relevant features from ECG signals\cite{b9}. Awni Y. Hannun compared the diagnostic accuracy of DNN models and cardiologists, and found that the classification sensitivity obtained by the ANN model was even better than that of experts\cite{b10}.

\subsection{Reducing the overhead of deep models}
Most of the related studies aim to improve the accuracy of heartbeat classification, but pay little attention to hardware resource consumption\cite{b11, b12}. Despite the superior performance of ANN models, one of the main obstacles to deploying DNNs on lightweight hardware is that they often require abundant computing and memory resources. With the increase in the application scenarios of wearable devices and mobile handheld devices, most of these devices have no enough memory and computing resources for the computation of DNN models. Reducing the memory overhead through compressing DNN models is urgently needed.

Current effective network compression methods mainly include Parameter Pruning\cite{b13, b14}, Parameter Quantization\cite{b15}, Low-Rank Decomposition\cite{b16, b17}, Knowledge Distillation\cite{b18, b19}, and Transferred/Compact Convolutional Filters\cite{b20}. Among the existing compression technologies, binarization methods have attracted much attention, in which the network parameters are represented by 1-bit instead of floating-point numbers, reducing the memory overhead greatly. BinaryNet\cite{b21} and XNOR-NET\cite{b22} both are well-established methods for binarization, achieving 32$\times$ memory savings on CPUs. Liu Z et al. proposed Bi-Real Net creatively, which can effectively reduce the information loss caused by binarization\cite{b23}. The ReActNet proposed by Liu Z et al. can introduce learnable parameters to optimize the activation function\cite{b24}, thereby increasing the amount of information carried by the activation value of the BNN.

\subsection{Quantization compression of CNN in ECG monitoring}
CNNs have been used in monitoring arrhythmias and quite a few quantitative compression methods are applied in reducing memory overhead. Li et al. \cite{b25} proposed a hierarchical quantization method based on greedy algorithm, and reached an accuracy of 95.39\% in the 17-categories test, which is only 0.33\% lower than that of the original network and reduces memory consumption by 15$\times$. Huang et al.\cite{b26} proposed the CNN model with incremental quantization aiming at classifying ECG signals in 17-categories, which achieved an accuracy of 92.76\% and a memory occupation of 39.34KB. The bCNN model\cite{b27} aimed at ECG binary-classification with binarized weight, and achieved an accuracy of 97.5\%.

Nevertheless, a search of the literature revealed few studies which concentrate on ECG classification to achieve an end-to-end binarized network, which is also the focus of our works.

\section{Methodology}\label{method}
In this section, we first introduce the structure of the baseline network and describe some of the means we utilize to binarize the network.

\begin{figure}[htbp]
\centerline{
\includegraphics[width=0.45\textwidth]{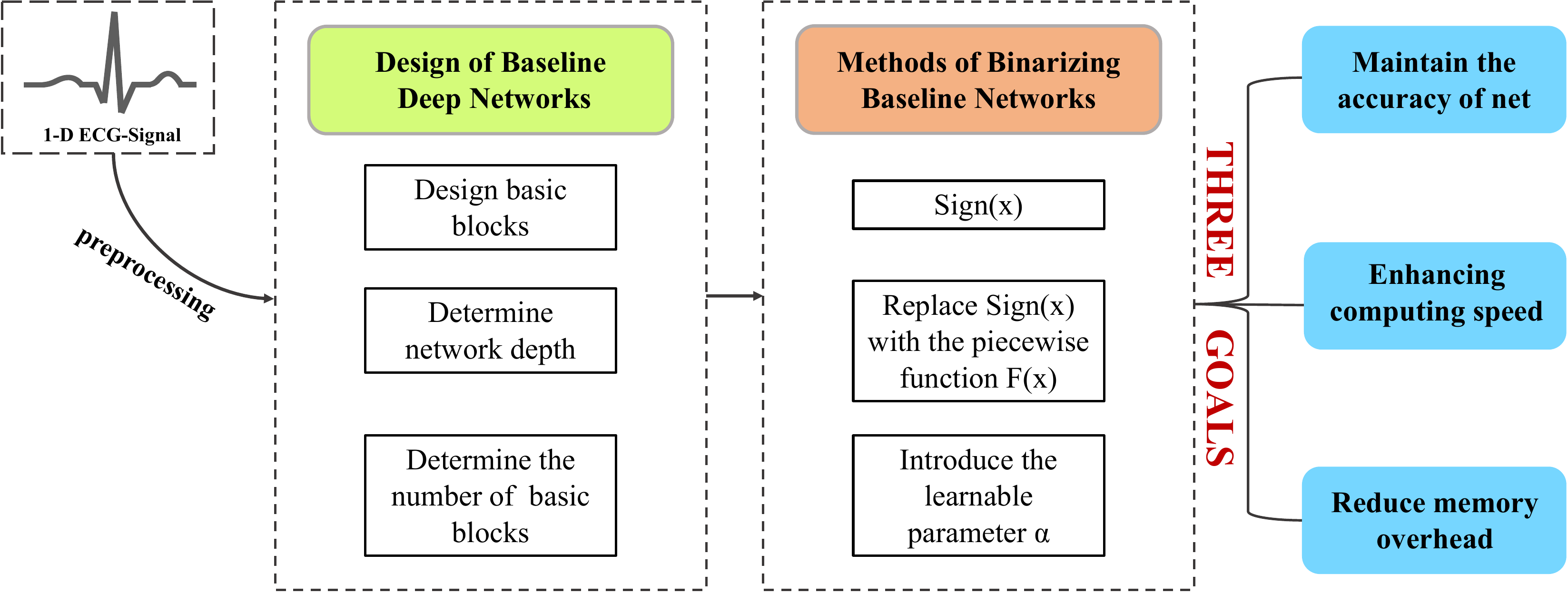}}
\caption{Overview of methods adopted}
\label{fig-arch}
\end{figure}

As shown in Fig. \ref{fig-arch}, the overall implementation process can be divided into two parts. Firstly, a full-precision baseline network was precisely designed. We conducted a lot of experiments to determine the depth of the network and the hyperparameters of each layer. Based on the architecture of the baseline network, we applied binarization methods to the weights and activations of the model, and obtained a binarized classifier.

\subsection{Design of the Baseline Deep Networks}
\begin{figure}[htbp]
\centerline{
\includegraphics[width=0.38\textwidth]{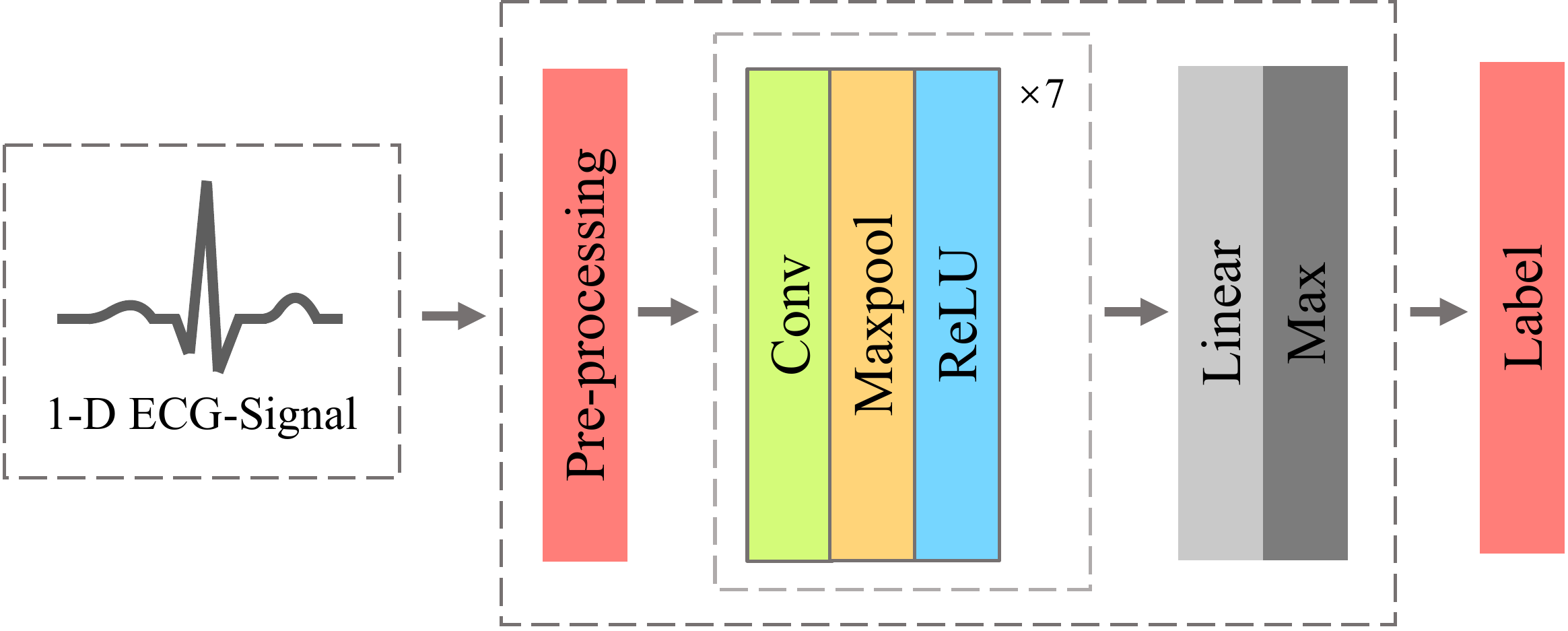}}
\caption{The architecture of the baseline network}
\label{fig-baseline}
\end{figure}

The architecture of the baseline arrhythmia classifier model is shown in Fig. \ref{fig-baseline}. The model is generally composed of seven convolutional basic blocks and a dense layer, in which the basic blocks are used for feature extraction while the dense layer is used for ECG classification. Each basic block contains a convolution layer and a pooling layer, and the ReLU activation function is placed between the basic blocks. For the classification task of five types of ECG signals, the input was the normalized long-term ECG signal, consisting of 3600 sampling points with a duration of 10s. The network depth should not be too small so that the learning effect of the network would be better.

\subsection{Methods of Binarization}\label{sec-bin}
\begin{figure}[htbp]
\centerline{
\includegraphics[width=0.35\textwidth]{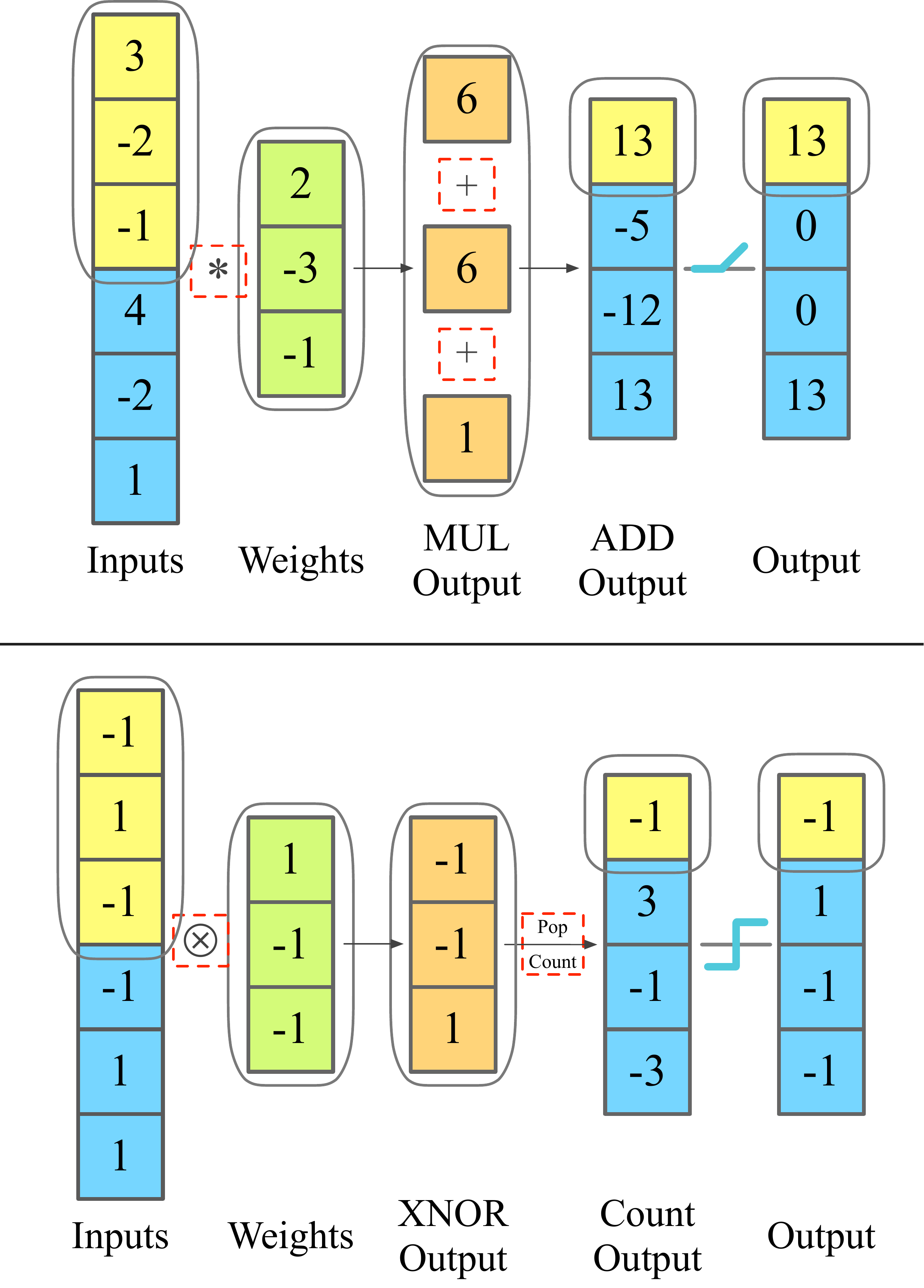}}
\caption{The respective operation methods of full-precision convolution and binary convolution. The top part is full-precision convolution, which includes floating-point multiplication and addition, taking up a lot of computational overhead; the bottom part is binary convolution, the multiplication operation is equivalent to the 1-bit operation XNOR, and the addition operation is replaced by counting.}
\label{fig-XNOR}
\end{figure}

The convolutional layers of a convolutional neural network contain lots of convolutional computations, and these operations are all floating-point multiplications and additions. The binarized convolutional network is to train the weight parameters of convolution kernels or activations to +1 or -1:
\begin{equation}
	{x}_{b}=Sign(x_{r})=\left\{
		\begin{aligned}
			& -1, & x < \alpha\\
			& +1, & x\geq \alpha
		\end{aligned}
		\right.
		\label{sign(x)}
\end{equation}
where $x$ represents the weights or activations, $b$ refers to the binarized weights or activations, and $r$ refers to real-values i.e. floating-point values. Through this form of conversion, the floating-point multiplication operation in the convolution can be replaced by the 1-bit operation XNOR, and the counting operation can achieve the effect of the floating-point addition operation, as depicted in Fig. \ref{fig-XNOR}.

The binarized network model would obtain a higher computing speed obviously, and its bit-widths should be extremely small. However, the problem of vanishing gradients that binarizing network brings should be taken into consideration. The derivative value in the back-propagation(BP) algorithm is obtained by the chain rule. Once the BP algorithm is used directly to process the network parameters which have been processed by the $Sign$ function, the resulting gradients are almost all 0, and there is no point in continuing to train the network. To solve the issue mentioned above, it is needed to preserve the gradients of real-values when training the binarized network. In the forward inference phase, we use $Sign(x)$ as the value of the weights or activations described by (\ref{sign(x)}), and let the gradient of $F(x)$ be the gradient of weights or activations during back-propagation. There are many options for the $F$ function here, such as $tanh$ and $hard-tanh$. In our study, we chose the $tanh$ function expressed in (\ref{tanh}), and the piecewise polynomial expressed in (\ref{Poly}). The gradient of the $tanh$ function and the polynomial can be obtained by (\ref{tanh-Grad}) and (\ref{Poly-Grad}), respectively.

\begin{equation}
	tanh(x)=\frac{e^{x} - e^{-x}}{e^{x} + e^{-x}}
	\label{tanh}
\end{equation}

\begin{equation}
	F(x)=\left\{
	\begin{aligned}
		& -1, & x<-1\\
		& 2x+x^{2}, & -1\leq x<0\\
		& 2x-x^{2}, & 0\leq x\leq1\\
		& +1, & x>1
	\end{aligned}
	\right.
	\label{Poly}
\end{equation}

\begin{equation}
	Grad(x_{b})=\frac{\partial tanh(x_{r})}{\partial x_{r}} = 1-tanh^{2}(x_{r})
	\label{tanh-Grad}
\end{equation}

\begin{equation}
	Grad(x_{b})=\frac{\partial F({x}_{r})}{\partial {x}_{r}}=\left\{	
	\begin{aligned}
		& 2+2{x}_{r}, & -1\leq {x}_{r}<0\\
		& 2-2{x}_{r}, & 0\leq {x}_{r} \leq 1\\
		& 0, & otherwise
	\end{aligned}
	\right.
	\label{Poly-Grad}	
\end{equation}

\begin{figure}[htbp]
\centerline{
\includegraphics[width=0.35\textwidth]{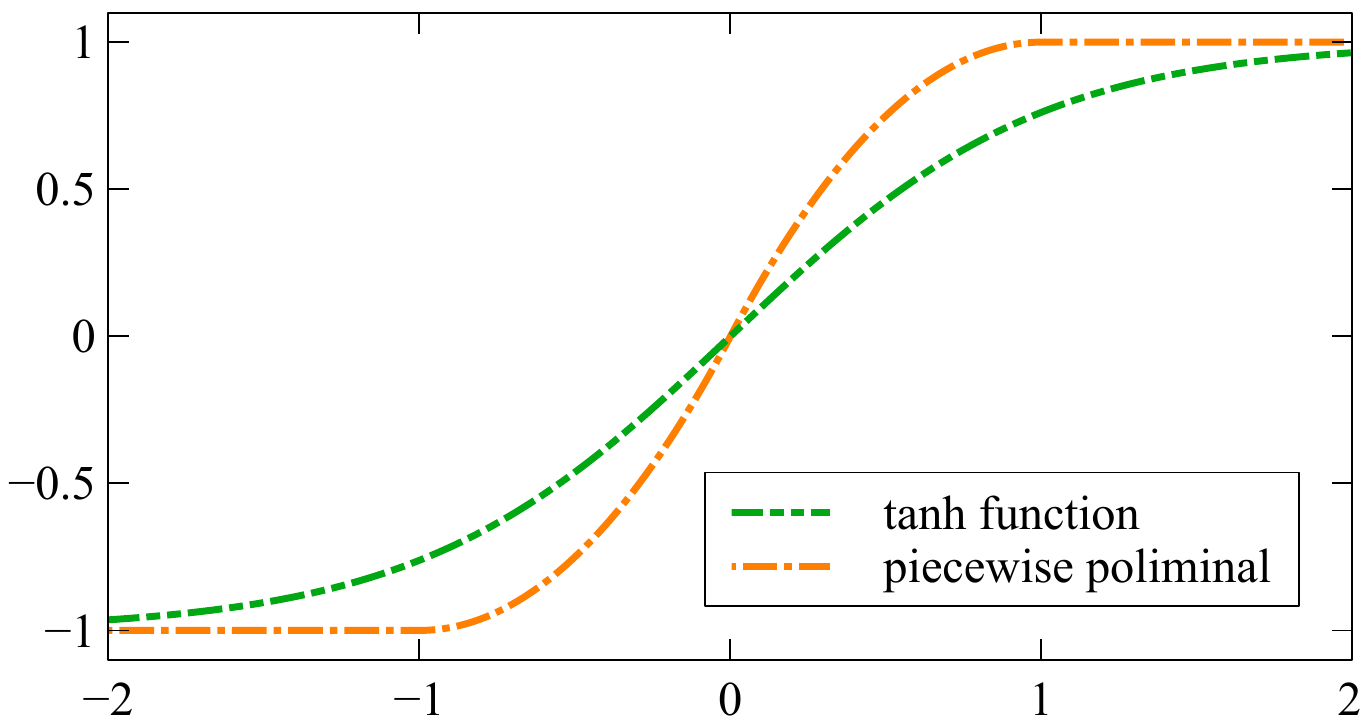}}
\caption{Tanh function and polynomial used. Piecewise polynomial would be better fitted to the Sign function}
\label{Binary-Conver}
\end{figure}

Compared with the $tanh$ function, the curve of equation (\ref{Poly}) fits the $Sign$ function better, as shown in Fig. \ref{Binary-Conver} .

As equation (\ref{sign(x)}) described, weights or activations are activated as +1 or -1 depending on their value relative to the threshold $\alpha$. A common practice is to set $\alpha$ to 0, meaning that weights or activations less than zero are assigned -1, otherwise +1. In our study, we try to set the threshold $\alpha$ as a learnable variable considering all weights or activations using the same threshold seems inappropriate. Each output channel corresponds to a learnable $\alpha$, including in the dense layer.

\section{Experiment}\label{exp}
In this section, we introduce the dataset used in the experiment and the implementation details of the baseline full-precision network firstly. Based on the baseline network model, we binarize the weights and activations to explore the effects of various binarization methods and explain the proposed topographic-based(TB) strategy for adjusting the learning rate when training binarized networks. Then we compare the accuracy of our binarized network model with the state-of-the-art models proposed by other researchers. Finally, we show the memory overhead and computation and analyze the shortcomings of the model and the points that can be improved. All the experiments have been performed using an Intel-i5 PC, with CPU speed of 2.3 GHz, 16 GB RAM and PyTorch 1.10.2.

\subsection{Dataset and Implementation of the Baseline Network}
The dataset used in the experiment comes from the Massachusetts Institute of Technology-Boston’s Beth Israel Hospital (MIT-BIH) Arrhythmia Database\cite{b28}. We extracted 7,740 ECG signal fragments with a duration of 10s, 80\% of which were selected as training data randomly, with the left 20\% as test data. All data were normalized in the experiment, and each signal fragment was categorized into five classes according to the Association for the Advancement of Medical Instrumentation (AAMI)\cite{b29}: ventricular ectopic (V), beat against other classes such as non-ectopic (N), supraventricular ectopic (S), fusion (F) and unknown (Q). The training-test splits for each category are shown in Table \ref{Dataset}.

\begin{table}[htbp]
\caption{The training-test splits for each category}
\begin{center}
\begin{tabular}{ccccc}
\hline
 & ALL & TRAIN & TEST & TEST RATIO\\
\hline
F & 100 & 85 & 15 & 0.15 \\
N & 5,186 & 4,149 & 1,037 & 0.2 \\
Q & 19 & 15 & 4 & 0.21 \\
S & 545 & 455 & 90 & 0.165 \\
V & 1,890 & 1,488 & 402 & 0.21 \\
Total & 7,740 & 6,192 & 1,548 & 0.2\\
\hline
\end{tabular}
\label{Dataset}
\end{center}
\end{table}

The baseline network mainly consists of 7 convolutional basic blocks and 1 dense layer. There will be a huge impact on the final result if bias is added to each layer even if it’s small in the binarized network. Aiming at constructing a model structure suitable for binarization, the biases of all convolutional layers and the dense layer are removed. Through extensive experiments, we obtained the structures shown in Table \ref{FPN}.

\begin{table}[htbp]
\caption{Architecture of the baseline network}
\begin{center}
\begin{tabular}{ccccc}
\hline
Label & Layer & Kernel Size & Stride\&Padding & Params Count\\
\hline
1 & Conv1D & 8 $\times$ 1 $\times$ 16 & 2, 7 & 128\\
2 & MaxPool & 8 & 4 & 0\\
3 & Conv1D & 12 $\times$ 8 $\times$ 12 & 2, 5 & 1,152\\
4 & MaxPool & 4 & 2 & 0\\
5 & Conv1D & 32 $\times$ 12 $\times$ 9 & 1, 4 & 3,456\\
6 & MaxPool & 5 & 2 & 0\\
7 & Conv1D & 64 $\times$ 32 $\times$ 7 & 1, 3 & 14,336\\
8 & MaxPool & 4 & 2 & 0\\
9 & Conv1D & 64 $\times$ 64 $\times$ 5 & 1, 2 & 20,480\\
10 & MaxPool & 2 & 2 & 0\\
11 & Conv1D &64 $\times$ 64 $\times$ 3 & 1, 1 & 12,288\\
12 & MaxPool & 2 & 2 & 0\\
13 & Conv1D & 72 $\times$ 64 $\times$ 3 & 1, 1 & 13,824\\
14 & MaxPool & 2 & 2 & 0\\
15 & Dense & 216 $\times$ 5 & {} & 1,080\\
\hline
\end{tabular}
\label{FPN}
\end{center}
\end{table}

We added a BatchNorm(BN) layer between the convolutional layer and the pooling layer for each convolutional basic block, and a dropout layer between the last basic block and the fully connected layer.

\begin{figure}[htbp]
\centerline{
\includegraphics[width=0.36\textwidth]{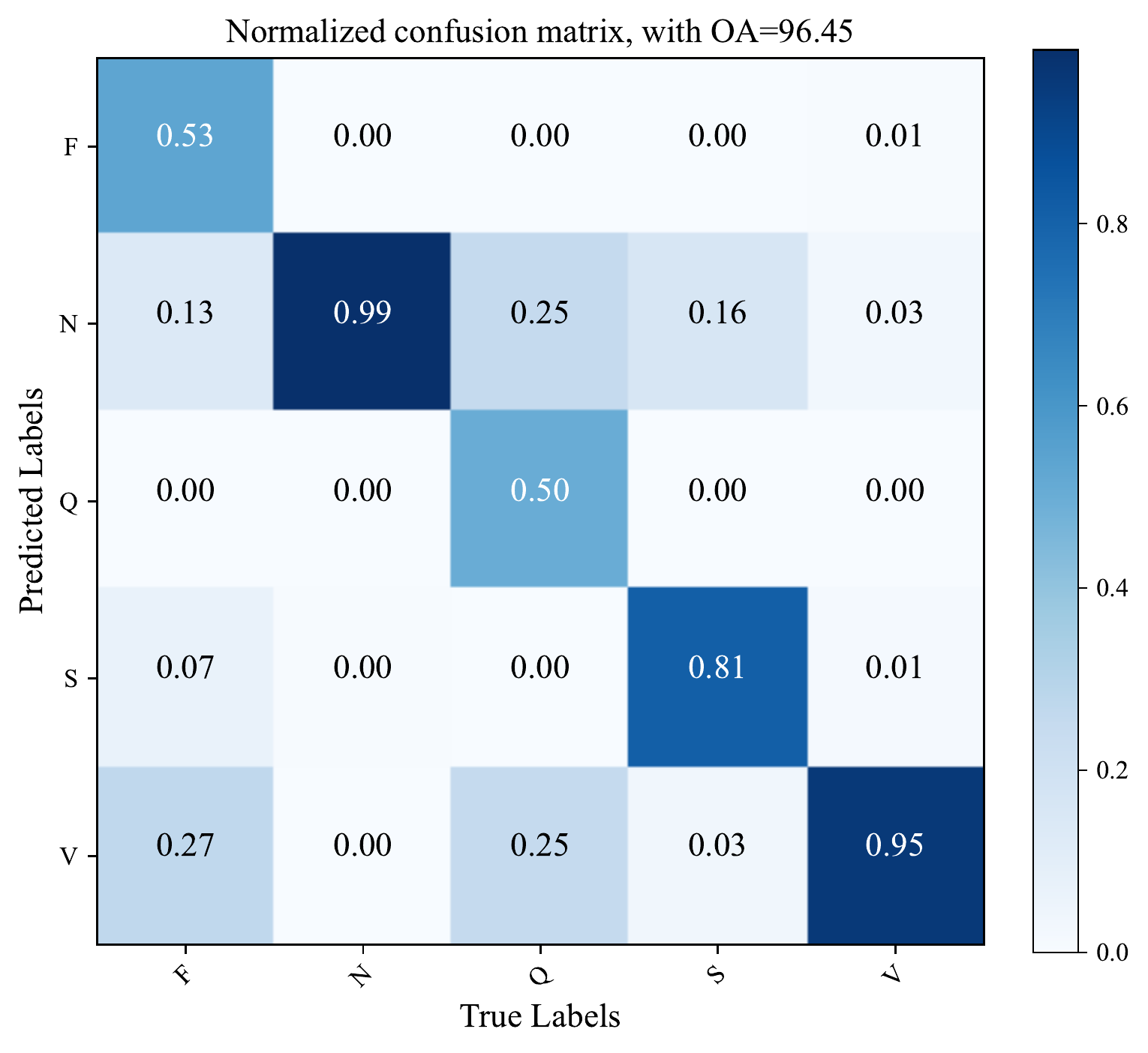}}
\caption{Confusion matrix of the Baseline Model}
\label{baseline-cfm}
\end{figure}

 There are 67,376 32-bit floating-point weight parameters in the baseline network in total, among which the required storage size is 263.1875KB, the overall accuracy(OA) of the model is 96.45\%, and the classification confusion matrix is shown in Fig. \ref{baseline-cfm}. It can be seen that the classification accuracy of categories N and V reached 99\% and 95\%, respectively, while the classification accuracy of the other three categories was lower, with too little data on those categories.

\subsection{Implementation of Binarized Classifier}
In this part, we implement the binarization method mentioned in \ref{sec-bin}, train the binarized network model, obtain different models through the combination of different binarization transformation functions, and select the best binarized model.

\begin{figure}[htbp]
\centerline{
\includegraphics[width=0.35\textwidth]{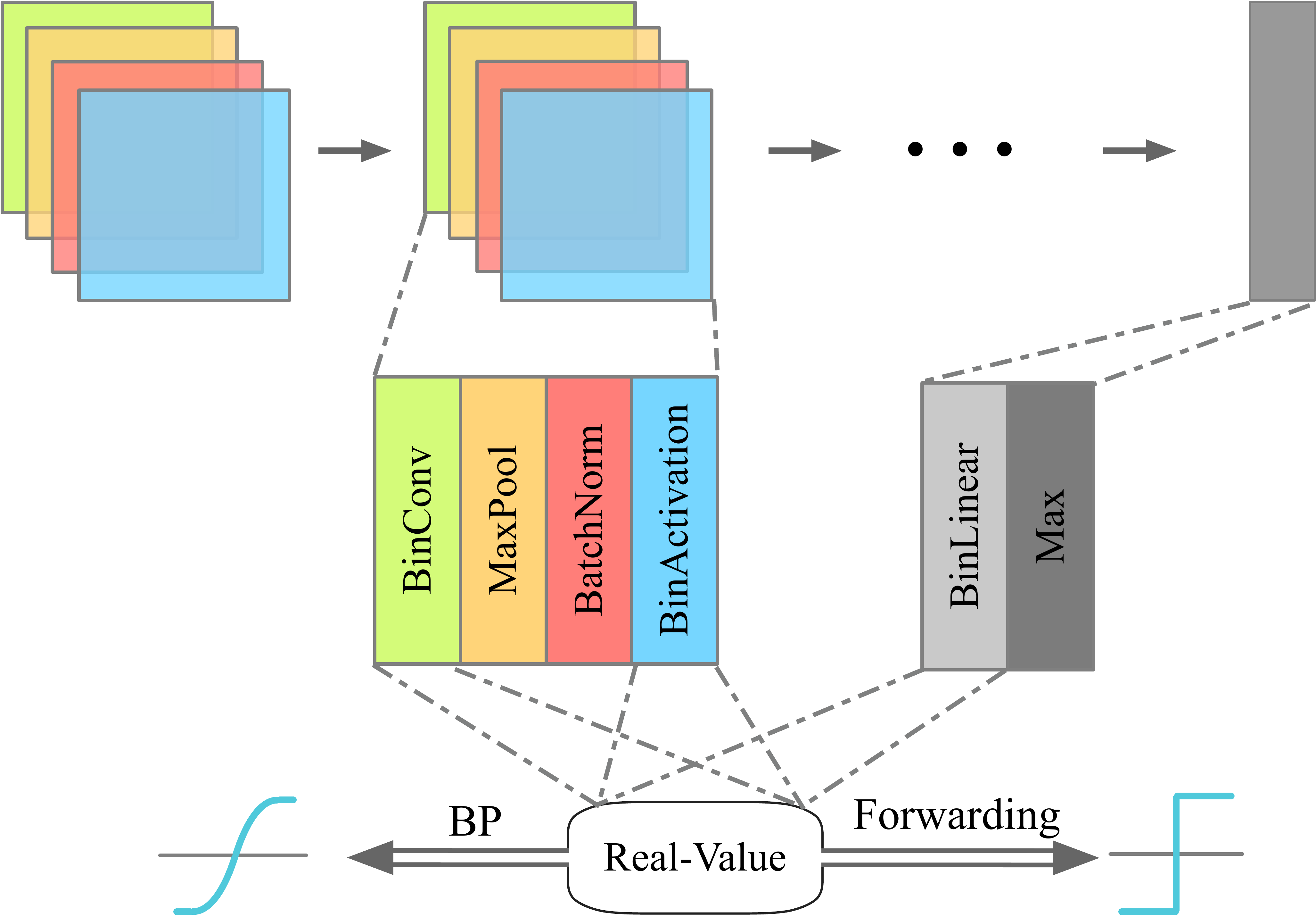}}
\caption{Diagram of the binarized model. The real-values of weights and activations will be preserved during training phase.}
\label{bin-ill}
\end{figure}

Compared with the baseline full-precision network, the binarized network comes with poor convergence. It’s clear in our experiments that the loss function won’t be able to converge during training, and the accuracy can’t improve with the BN layer being placed between the convolutional layer and the pooling layer. Only if the BN layer is placed after the pooling layer would there be a different result, thus we fine-tune the position of the BN layer in the architecture of the baseline network. In the binarized network model, the BN layer was placed between the pooling layer and activation layer in each binarized convolutional basic block, as illustrated in Fig. \ref{bin-ill}. Since activations and weights are binarized, it is equivalent to adding random noise, with regularization and suppression of overfitting.

Considering the attractiveness of bit-operations instead of floating-point operations, we binarized both the weights and activations.

\textbf{Training the BTTN model.} The threshold $\alpha$ in (\ref{sign(x)}) was set to be 0 firstly, with the gradient of $tanh$ being the gradient of both the weight and the activation as described by (\ref{tanh-Grad}). In this way, we obtained the binary network model(BTTN) with an accuracy of 94.90\%. The gradients of weights and activations could be calculated by $tanh(weight)$ and $tanh(activation)$ respectively.

\textbf{Training the BTPN model.} Compared with the $tanh$ function, the curve of the piecewise polynomial $F$ in (\ref{Poly}) is more fitted to the $Sign$ function, and its gradient should be more approximate. The threshold $\alpha$ in (\ref{sign(x)}) was still set to be 0, as before. We took the gradient of $tanh(weight)$ as the gradient of the weight and the gradient of $F(activation)$ as the gradient of the activation. The BTPN model with an accuracy of 95.67\% was obtained after training.

\textbf{Training the BPPN and BPTN model.} Since BTPN with piecewise polynomial applied achieves higher accuracy, the gradient of piecewise polynomial seems to be more suitable as the gradient for binarization. By further using the gradient of $F(weight)$ as the gradient of the weight, a BPPN model with an accuracy of 95.12\% was obtained after completed training. Similarly, we also trained a BPTN model with an accuracy of 94.83\% and the accuracy of these two models was lower than that of the BTPN model.

\textbf{Training the BTPN-$\alpha$ model.} We then tried to set the activation threshold $\alpha$ of each output channel as a respective learnable parameter, and the base model for this step was BTPN. Counterintuitively, the accuracy of the obtained BTPN-$\alpha$ model was only 94.96\%.

\begin{figure}[htbp]
\centerline{
\includegraphics[width=0.36\textwidth]{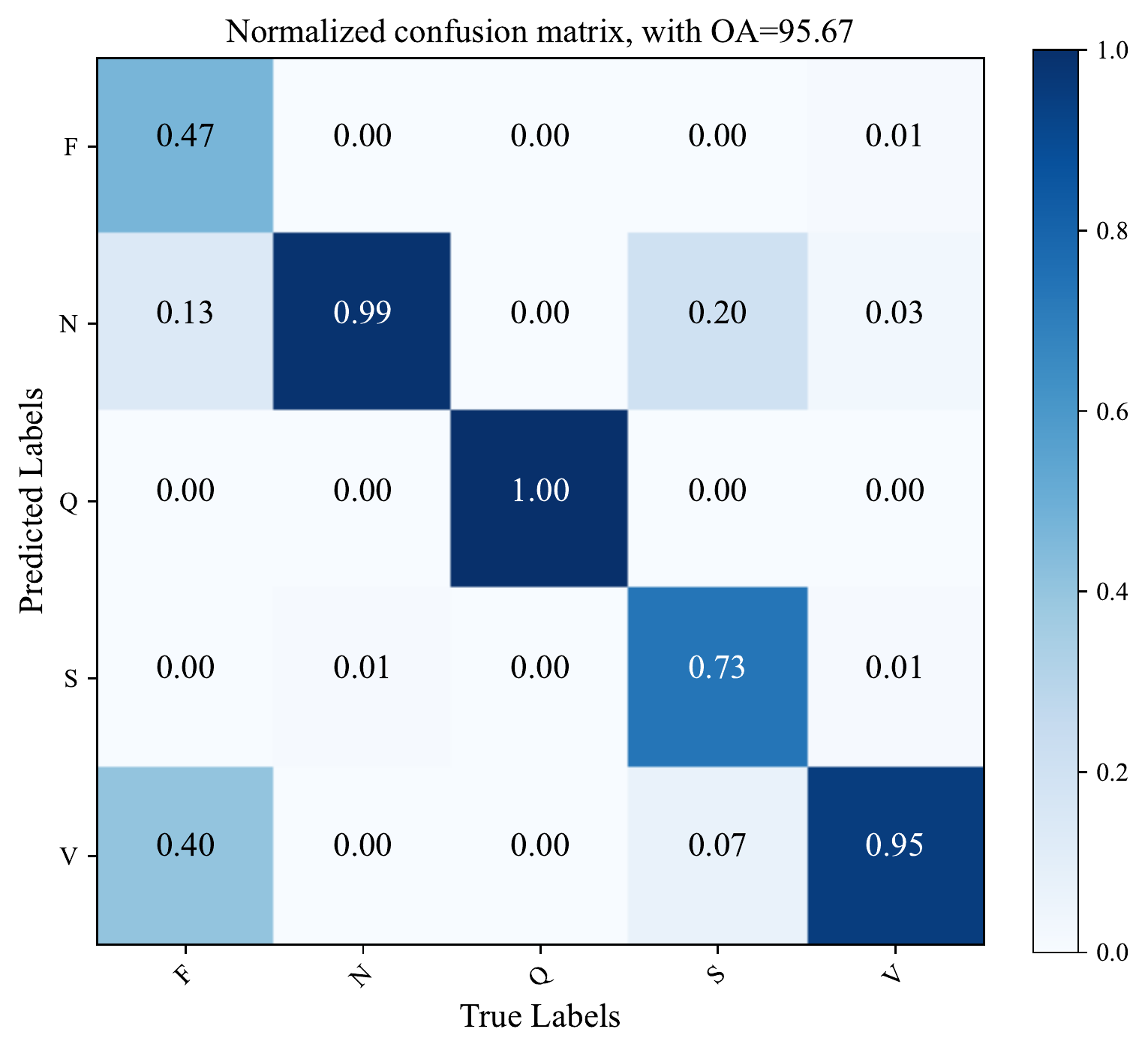}}
\caption{Confusion matrix of the proposed BPTN model}
\label{btpn-cfm}
\end{figure}

Finally, we got the BTPN model with an accuracy of \textbf{95.67\%}, which is only 0.78\% lower than that of the baseline network. The classification confusion matrix for each category is shown in Fig. \ref{btpn-cfm}.

\begin{figure}[htbp]
\centerline{
\includegraphics[width=0.4\textwidth]{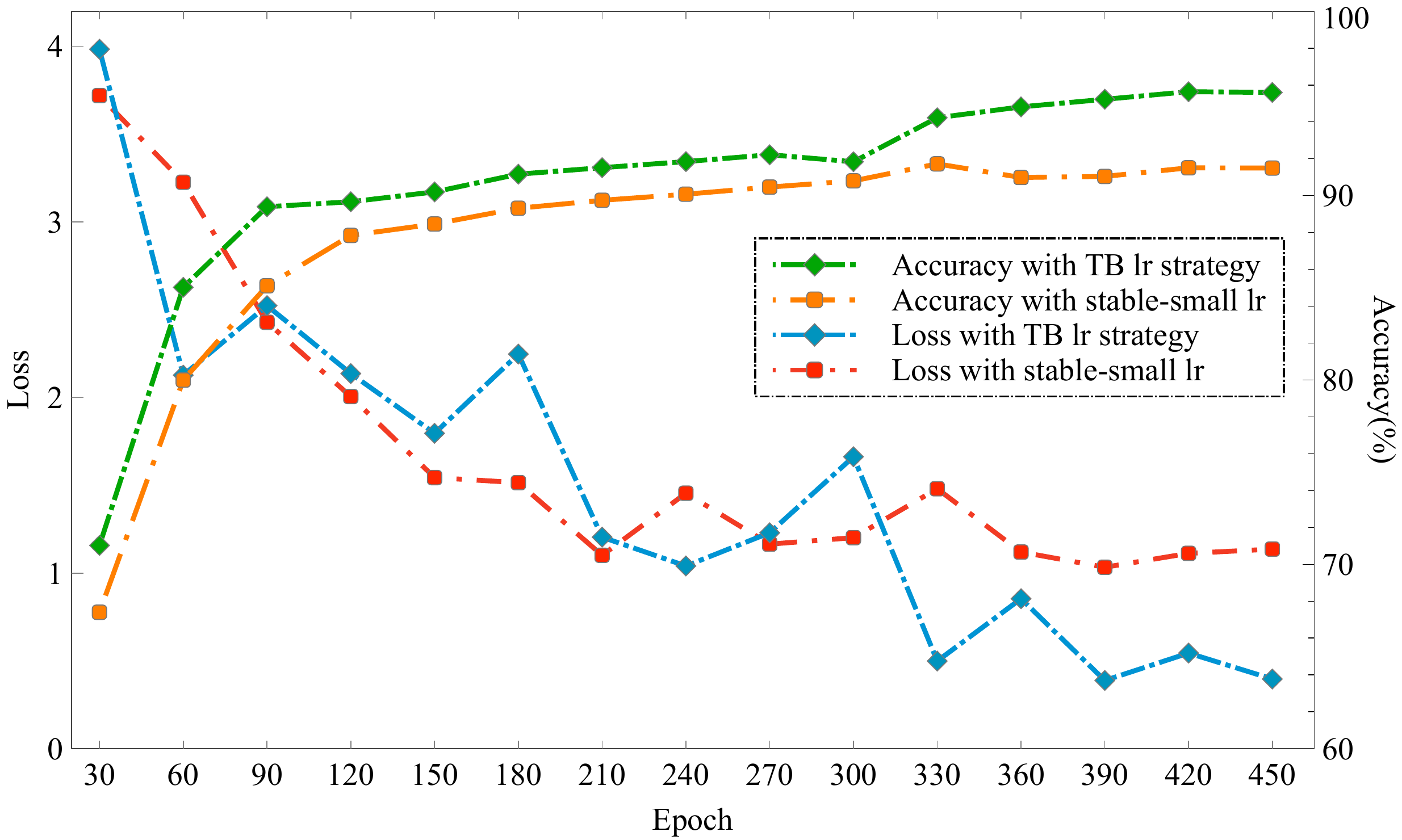}}
\caption{Accuracy and Loss. The strategy used for the learning rate has a great impact on the loss and accuracy of the binarized model.}
\label{Loss-acc}
\end{figure}

\begin{figure}[htbp]
\centerline{
\includegraphics[width=0.33\textwidth]{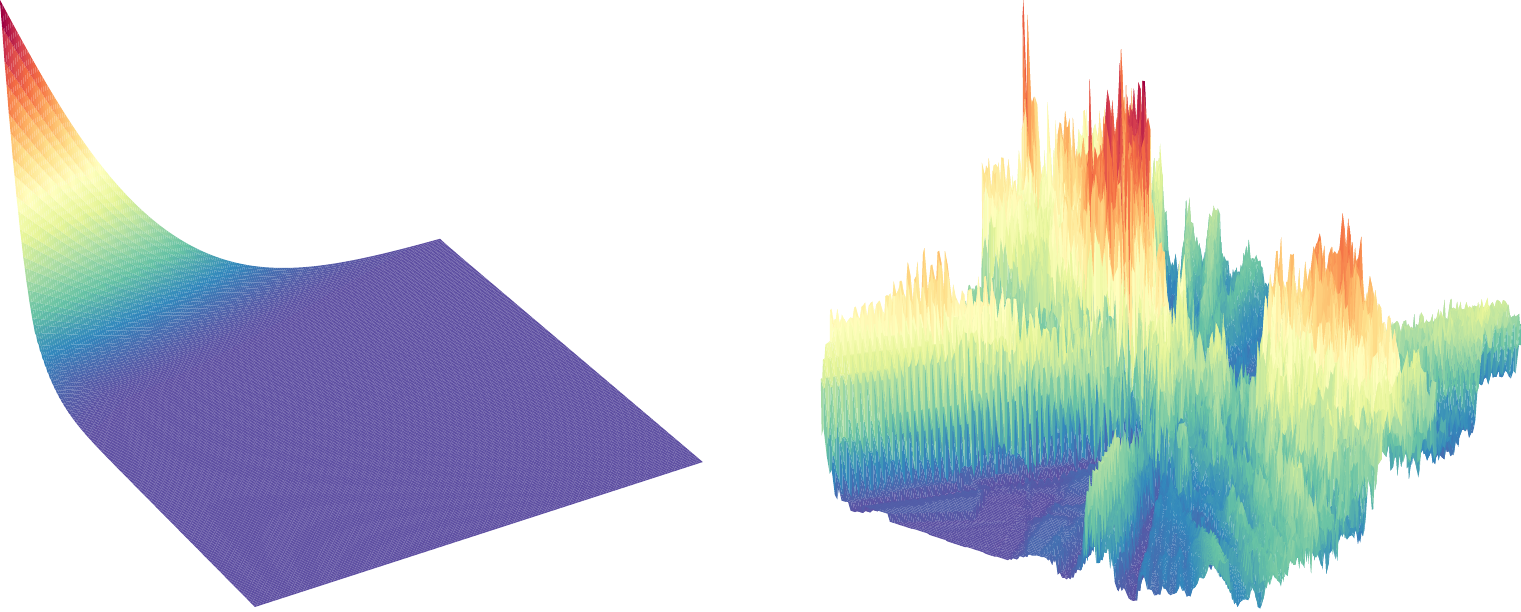}}
\caption{Loss-Landscape of baseline full-precision network and binarized network. The loss surface of the baseline full-precision network is quite smooth, while that of the binarized network model is undulating.}
\label{Loss-Land}
\end{figure}

Here we discuss the TB strategy for adjusting learning rate. After extensive experimentation, we found that if the learning rate throughout the training process is set to a relatively small value, the loss of the model will always remain high, and the accuracy will not be improved which could be observed in Fig. \ref{Loss-acc}. Based on the method proposed by Li, H et al.\cite{b30}, we plotted the loss topography of the baseline network versus the binarized network. As Fig. \ref{Loss-Land} presented, the loss surface of the baseline full-precision network is quite smooth, and the loss could easily decrease, while that of the binarized network model is undulating. If the learning rate is set too small, the loss during training is likely to fall into a local minimum value, so that the global optimal value cannot be further learned, and the accuracy cannot be improved. If the learning rate is set relatively large in the early stage of training, when the loss is relatively low and the accuracy rate is relatively high, we can reduce the learning rate and keep the loss and accuracy relatively stable. With this strategy,  binarized models with higher accuracy could be obtained.

\subsection{Accuracy comparison with other state-of-the-art models}
After precise full-precision network design and extensive experiments on binarization, a Full-Binarized Network(proposed BTPN model) with an OA of 95.67\% was obtained. For the sake of objectively evaluating the performance of the network model, comparing the proposed BTPN model with the state-of-the-art networks of other excellent researchers is of necessity. To this end, we conducted a comparative study with three methods: TBME2015\cite{b31}, ITBCS2019\cite{b32} and ISCAS2021\cite{b27}. Since the AAMI indicator recommends using the V and S categories to evaluate the performance, we chose SEN and PPR indicators of the V-category and overall accuracy to compare the performance of each network.

\begin{table}[htbp]
\caption{Comparison with other state-of-the-art models on overall accuracy and V category metrics}
\begin{center}
\begin{tabular}{cccccc}
\hline
Model & SEN(\%) & PPR(\%) & OA(\%) & Method$^{\mathrm{a}}$ & Sorts\\
\hline
TBME2015\cite{b29} & 93.9 & 90.6 & 96.5 & FP & 5\\
ITBCS2019\cite{b30} & 91.8 & 95.3 & 98.4 & FP & 5\\
ISCAS2021\cite{b31} & 80.6 & 90.9 & 96.8 & FB & 2\\
\hline
Baseline Model & 95.3 & 97.0 & 96.45 & FP & 5\\
BTPN(Proposed) & \textbf{94.8} & \textbf{96.2} & \textbf{95.67} & FB & 5\\ 
\hline
\multicolumn{6}{l}{$^{\mathrm{a}}$FP refers to Full-Precision, FB refers to Full-Binarized.}
\end{tabular}
\label{ACC-Com}
\end{center}
\end{table}

The results of the comparison are shown in Table \ref{ACC-Com}. The network architecture of TBME2015 and ITBCS2019 is CNN+MLP, while ISCAS2021, baseline network and BTPN are all CNN-based. What should be noted is that the BTPN model and the above network architectures have different classification categories for ECG signals with diverse processing methods for weights and activations at the same time, thus these two factors should be taken into consideration for fairness when comparing. Despite the fact that the OA of the proposed BTPN is 2.73\% lower than that of the full-precision model ITBCS2019 with the highest OA, most of the floating-point operations could be replaced by bit-operations(BOPs) since activations and weights are all 1-bit, reducing the memory overhead and improving the operation speed to a great extent. Compared with ISCAS2021’s binarized network, the proposed BTPN network achieves classification for more categories, and even outperforms ISCAS2021’s BNN on V-category, with OA only 1.13\% lower than it. The comparison clearly indicates that our BTPN has considerable accuracy while greatly reducing the memory and computational overhead, making it more suitable for real-time ECG monitoring and deployment on resource-constrained devices.

\subsection{Analysis and discussion}
Even if weights in each convolution layer and activations are all 1-bit, the convolution and pooling operations of the first basic block still need to participate in floating-point operations instead of bit operations at runtime for the input ECG signals are floating-point values (or integers, here we use floating-point numbers for analysis). In addition, each BN layer still needs to be stored and operated with floating-point numbers (Floating-point operations in BN layers could be replaced by operations such as shifting\cite{b22}, the calculation amount of BN layers only accounts for 1/300 of the whole, so here we still consider the operation of BN layers as floating-point operations, and the parameters are stored as floating-point numbers). Most CPUs of modern PCs are 64-bit and can perform 64 bit-operations in parallel while processors are practically 32-bit and can only perform 32-bit operations on watches, medical equipments and resource-constrained devices. The floating-point operations are uniformly calculated with 32-bit floating-point numbers, that is, the time to perform one 32-bit floating-point operation is roughly the same as the time to perform 32 bit-operations.

\begin{table}[htbp]
\caption{Comparison of memory overhead and computation between the proposed BTPN model and the baseline model}
\begin{center}
\begin{tabular}{cccc}
\hline
{} & Storage & Runtime Mem & OPs\\
\hline
Baseline FPN & 263.12KB & 444.93KB & $4.875\times 10^{6}$ FLOPs\\
BTPN & 10.62KB & 117.70KB & $2.458\times10^{5}$ FLOPs\\
\hline
Saving/Speedup & \textbf{24.8$\times$} & \textbf{3.78$\times$} & \textbf{12.65$\times$}$^{\mathrm{a}}$\\
\hline
\multicolumn{4}{l}{$^{\mathrm{a}}$We also took $4.471\times10^{6}$ BOPs in BTPN into consideration.}
\end{tabular}
\label{Perfo}
\end{center}
\end{table}

The results of calculations are shown in Table \ref{Perfo}. It can be seen that compared to the baseline model, the storage footprint of the proposed BTPN model is reduced by \textbf{24.8$\times$}, the memory overhead at runtime is reduced by \textbf{3.78$\times$} and the theoretical calculation speed is \textbf{12.65$\times$} faster while the accuracy loss is only 0.78\%, which can be said to be a cost-effective trade-off.

Overall, the binarized model greatly compresses the storage footprint and increases the computing speed, while retaining a considerable and reliable accuracy, enabling the deployment of real-time ECG monitoring on resource-constrained wearable devices. Nonetheless, based on the theory of binary computing, our classifier still has tremendous prospects for improvement. The proposed model still has many floating-point operations involved in the calculation, such as operations in the first convolution basic block and each BN layer. The storage overhead of the model has not been reduced to an extreme $32\times$, nor has the computational speed increased to the theoretical performance bottleneck. We will strive to solve the drawbacks mentioned above, so that the model has a higher computing speed, occupies a smaller runtime memory, and is more suitable for real-time monitoring.

\section{Conclusion}\label{conclu}
In this work, we present an efficient convolutional neural network to recognize 1-D long-term ECG signal segments, dubbed BTPN. Based on the architecture of proposed baseline full-precision network, we adopted the binarization method, thereby replacing most of the floating-point operations with 1-bit operations. It maintains considerable accuracy while increasing computational speed and reducing resource overhead. The classification accuracy of the proposed BTPN model in the MIT-BIH Arrhythmia Database reaches 95.67\% in a five-class test. In the future, we will consider the processing of the input signal and the optimization of BatchNorm layers to further improve the computing performance and runtime memory compression ratio, and implement this work on a hardware platform for real-time ECG monitoring.


\begin{thebibliography}{00}
\bibitem{b1} M. A. Serhani, A. N. Navaz, H. Al Ashwal, and N. Al Qirim, ``Ecg-based arrhythmia classification \& amp; clinical suggestions: An incremental approach of hyperparameter tuning," in Proceedings of the 13th International Conference on Intelligent Systems: Theories and Applications, 2020. [Online]. Available: doi.org/10.1145/3419604.3419787
\bibitem{b2} K. Simonyan and A. Zisserman, ``Very deep convolutional networks for large-scale image recognition," arXiv preprint arXiv:1409.1556, 2015.
\bibitem{b3} Sayad A.T., Halkarnikar P.P., ``Diagnosis of heart disease using neural network approach," International Journal of Advances in Science Engineering and Technology, vol. 2, no. 3, pp. 88-92, 2014.
\bibitem{b4} M. Llamedo and J. P. Martinez, ``Heartbeat classification using feature selection driven by database generalization criteria," IEEE Transactions on Biomedical Engineering, vol. 58, no. 3, pp. 616-625, Mar. 2011.
\bibitem{b5} K. Park, B. Cho, D. Lee, S. Song, J. Lee, Y. Chee, I. Kim, and S. Kim, ``Hierarchical support vector machine based heartbeat classification using higher order statistics and hermite basis function," in 2008 Computers in Cardiology, 2008, pp. 229-232.
\bibitem{b6} S. Osowski, L. Hoai, and T. Markiewicz, ``Supportvectormachine-based expert system for reliable heartbeat recognition," IEEE Transactions on Biomedical Engineering, vol. 51, no. 4, pp. 582-589, Apr. 2004.
\bibitem{b7} C. Ye, M. T. Coimbra, and B. Vijaya Kumar, ``Arrhythmia detection and classification using morphological and dynamic features of ecg signals," in 2010 Annual International Conference of the IEEE Engineering in Medicine and Biology, 2010, pp. 1918-1921.
\bibitem{b8} G. D. Clifford, C. Liu, B. Moody, L.-w. H. Lehman, I. Silva, Q. Li, A. E. Johnson, and R. G. Mark, ``AF classification from a short single lead ecg recording: The physionet/computing in cardiology challenge 2017," in 2017 Computing in Cardiology (CinC), 2017, pp. 1-4.
\bibitem{b9} X. Fan, Q. Yao, Y. Cai, F. Miao, F. Sun, and Y. Li, ``Multiscaled fusion of deep convolutional neural networks for screening atrial fibrillation from single lead short ecg recordings," IEEE Journal of Biomedical and Health Informatics, vol. 22, no. 6, pp. 1744-1753, Nov. 2018.
\bibitem{b10} A. Y. Hannun, P. Rajpurkar, M. Haghpanahi, G. H. Tison, C. Bourn, M. P. Turakhia, and A. Y. Ng, ``Cardiologist-level arrhythmia detection and classification in ambulatory electrocardiograms using a deep neural network," Nature Medicine, vol. 25, no. 1, pp. 65-69, 2019. [Online]. Available: doi.org/10.1038/s41591-018-0268-3
\bibitem{b11} Y. Li, Y. Pang, J. Wang, and X. Li, ``Patient-specific ecg classification by deeper cnn from generic to dedicated," Neurocomputing, vol. 314, pp. 336-346, 2018.
\bibitem{b12} S. Zhou and B. Tan, ``Electrocardiogram soft computing using hybrid deep learning cnn-elm," Applied Soft Computing, vol. 86, p. 105778, 2020.
\bibitem{b13} S. Han, J. Pool, J. Tran, and W. J. Dally, ``Learning both weights and connections for efficient neural networks," in Proceedings of the 28th International Conference on Neural Information Processing Systems, 2015, pp. 1135-1143.
\bibitem{b14} Y. He, X. Zhang, and J. Sun, ``Channel pruning for accelerating very deep neural networks," in 2017 IEEE International Conference on Computer Vision (ICCV), 2017, pp. 1398-1406.
\bibitem{b15} J. Wu, C. Leng, Y. Wang, Q. Hu, and J. Cheng, ``Quantized convolutional neural networks for mobile devices," in 2016 IEEE Conference on Computer Vision and Pattern Recognition (CVPR), 2016, pp. 4820- 4828.
\bibitem{b16} L. Vadim, G. Yaroslav, R. Maksim, V. O. Ivan, and S. L. Victor, ``Speeding-up convolutional neural networks using fine-tuned cp-decomposition," in 3rd International Conference on Learning Representations, May 2015, Conference Track Proceedings, 2015.
\bibitem{b17} M. Jaderberg, A. Vedaldi, and A. Zisserman, ``Speeding up convolutional neural networks with low rank expansions," in Proceedings of the British Machine Vision Conference, BMVA Press, 2014.
\bibitem{b18} Z. Xu, Y.-C. Hsu, and J. Huang, ``Training shallow and thin networks for acceleration via knowledge distillation with conditional adversarial networks," in ICLR, 2018.
\bibitem{b19} Y. Chen, N. Wang, and Z. Zhang, ``Darkrank: Accelerating deep metric learning via cross sample similarities transfer," in Proceedings of the Thirty-Second AAAI Conference on Artificial Intelligence, 2018, pp. 2852-2859.
\bibitem{b20} M. Sandler, A. Howard, M. Zhu, A. Zhmoginov, and L.-C. Chen, ``Mobilenetv2: Inverted residuals and linear bottlenecks," in 2018 IEEE/CVF Conference on Computer Vision and Pattern Recognition, 2018, pp. 4510-4520.
\bibitem{b21} M. Courbariaux and Y. Bengio, ``Binarynet: Training deep neural networks with weights and activations constrained to +1 or -1," in CoRR, arXiv preprint arXiv: 1602.02830, 2016.
\bibitem{b22} M. Rastegari, V. Ordonez, J. Redmon, and A. Farhadi, ``Xnor-net: Imagenet classification using binary convolutional neural networks," in ECCV (4), 2016, pp. 525-542.
\bibitem{b23} Z. Liu, B. Wu, W. Luo, X. Yang, W. Liu, and K.-T. Cheng, ``Bi-real net: Enhancing the performance of 1-bit cnns with improved representational capability and advanced training algorithm," in Computer Vision - ECCV 2018, 2018, pp. 747-763.
\bibitem{b24} Z. Liu, Z. Shen, M. Savvides, and K.-T. Cheng, ``Reactnet: Towards precise binary neural network with generalized activation functions," in Computer Vision - ECCV 2020, 2020, pp. 143-159.
\bibitem{b25} Z. Li, H. Li, X. Fan, F. Chu, S. Lu, and H. Liu, ``Arrhythmia classifier using a layer-wise quantized convolutional neural network for resource-constrained devices," in Proceedings of the 2020 International Symposium on Artificial Intelligence in Medical Sciences, 2020, pp. 38-44.
\bibitem{b26} J. Huang, Z. Liu, and H. Liu, ``An efficient arrhythmia classifier using convolutional neural network with incremental quantification," Journal of Physics: Conference Series, vol. 1966, no. 1, pp. 012022, Jul. 2021.
\bibitem{b27} D. L. T. Wong, Y. Li, D. John, W. K. Ho, and C. H. Heng, ``Resource and energy efficient implementation of ecg classifier using binarized cnn for edge ai devices," in 2021 IEEE International Symposium on Circuits and Systems (ISCAS), 2021, pp. 1-5.
\bibitem{b28} A. L. Goldberger, L. A. N. Amaral, L. Glass, J. M. Hausdorff, P. C. Ivanov, R. G. Mark, J. E. Mietus, G. B. Moody, C.-K. Peng, and H. E. Stanley, ``Physiobank, physiotoolkit, and physionet," Circulation, vol. 101, no. 23, pp. e215-e220, 2000.
\bibitem{b29} ``Testing and reporting performance results of cardiac rhythm and ST segment measurement algorithms," Association for the Advancement of Medical Instrumentation, Arlington, VA, 1987.
\bibitem{b30} H. Li, Z. Xu, G. Taylor, C. Studer, and T. Goldstein, ``Visualizing the loss landscape of neural nets," in Proceedings of the 32nd International Conference on Neural Information Processing Systems, 2018, p. 6391-6401.
\bibitem{b31} S. Kiranyaz, T. Ince, and M. Gabbouj, ``Real-time patient-specific ecg classification by 1-d convolutional neural networks," IEEE Transactions on Biomedical Engineering, vol. 63, no. 3, pp. 664-675, Mar. 2016.
\bibitem{b32} N. Wang, J. Zhou, G. Dai, J. Huang, and Y. Xie, ``Energy-efficient intelligent ecg monitoring for wearable devices," IEEE Transactions on Biomedical Circuits and Systems, vol. 13, no. 5, pp. 1112-1121, Oct. 2019.
\end{thebibliography}
\end{document}